\documentclass[11pt]{article}
\usepackage{acl-hlt2011}
\usepackage{times}
\usepackage{latexsym}
\usepackage{amsmath}
\usepackage{multirow}
\usepackage{url}
\usepackage{graphicx}
\usepackage{enumerate}

\title{CIS at TAC Cold Start 2015: \\Neural Networks and Coreference Resolution for Slot Filling}

\author{Heike Adel and Hinrich Sch\"{u}tze \\
  Center for Information and Language Processing (CIS)\\
  LMU Munich \\
  Germany\\
  {\tt heike.adel@cis.lmu.de}}

\date{}

\begin{document}
\maketitle
\begin{abstract}
This paper describes the CIS slot filling system for the TAC Cold Start
evaluations 2015.
It extends and improves the system we have built for the
evaluation last year. This paper mainly
describes the changes to our last year's system.
Especially, it focuses on the coreference and classification
component. For coreference, we have performed several analysis
and prepared a resource to simplify our end-to-end
system and improve its runtime. For classification, 
we propose to use neural networks.
We have trained convolutional and recurrent neural networks
and combined them with traditional evaluation methods, 
namely patterns and support vector machines.
Our runs for the 2015 evaluation have been designed
to directly assess the effect of each network
on the end-to-end performance of the system.
The CIS system achieved rank 3 of all slot filling
systems participating in the task.
\end{abstract}

\section{Introduction}
The TAC KBP Slot Filling task addresses the challenge of gathering information
about entities (persons, organizations or geo-political entities) 
from a large amount of unstructured text data. 
Previous evaluations showed that this task includes a variety
of challenges like document retrieval, coreference resolution, location inference, 
cross-document inference and relation extraction / classification.
In our slot filling system, we address most of these challenges 
(except for cross-document inference which we only consider
in the context of location inference).
This paper focuses on the changes of our system compared
to last year \cite{cis2014}, especially on our relation classification
and coreference component.
We propose to tackle relation classification with neural networks and
show the importance of coreference resolution for slot filling.
Additional changes which led to significant system improvements
included extension and automatic selection of training data and
genre specific processing of documents.

The remainder of the paper is organized as follows: First, an overview of
the slot filling system is presented (Section \ref{sec:overview}). 
Second, the changes of the different components of the
system are described in detail. The forth Section describes how we integrated coreference
resolution and Section \ref{neuralNets} presents our 
neural classification models. Finally, the performance of the system 
in the shared task is presented.

\section{System overview}
\label{sec:overview}
\begin{figure*}
\centering
\includegraphics[width=0.7\textwidth]{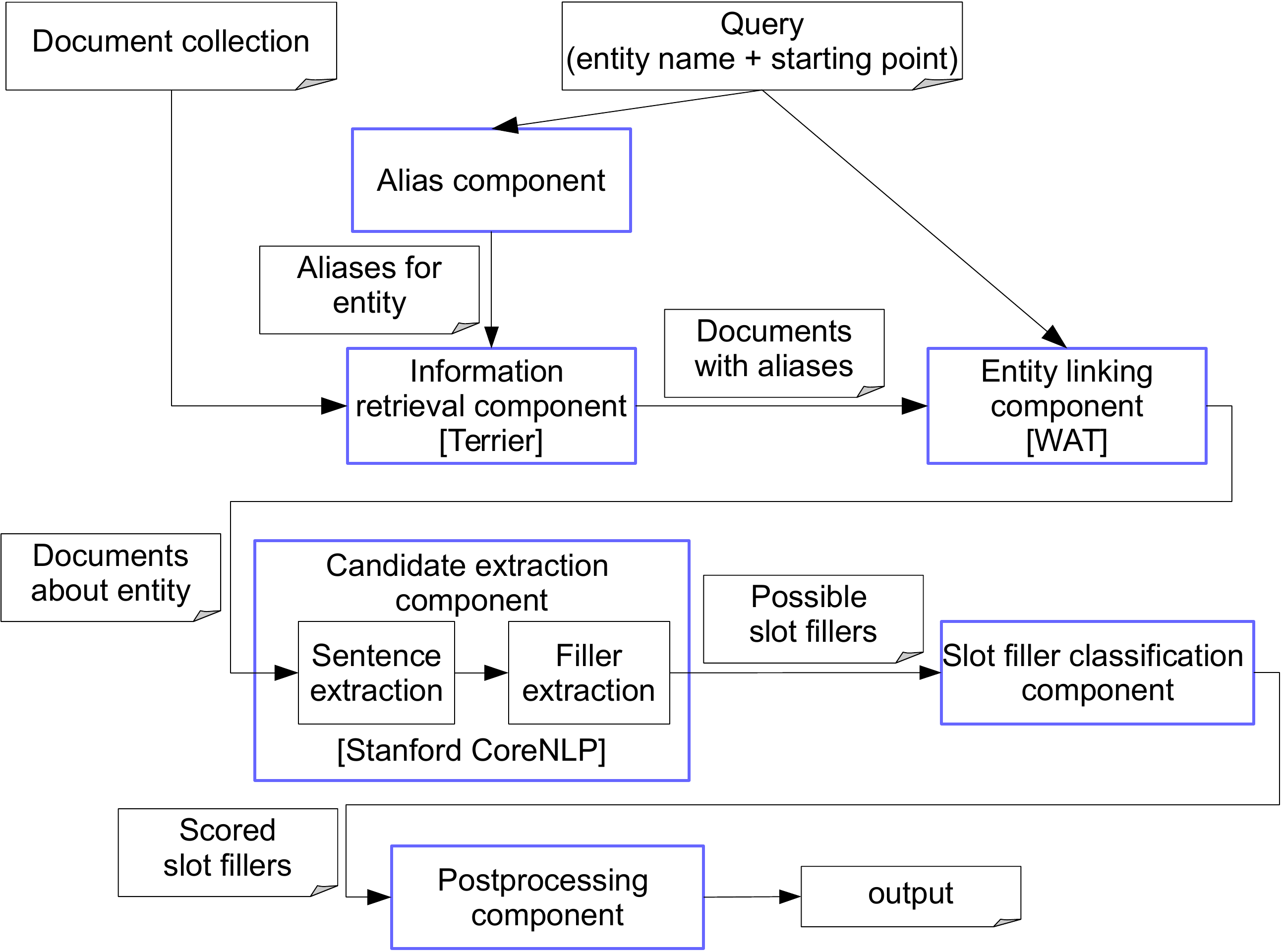}
\caption{System overview: Basic components of the CIS slot filling system}
\label{systemoverview}
\end{figure*}

Our slot filling system is an extension of our system from last year.
It addresses the slot filling task in a modular way. This has several advantages, including 
extensibility, componentwise analyzability and modular development.
Figure~\ref{systemoverview} shows the components of our system.
In order to gather information about a person, organization or geo-political entity, 
the following steps need to be performed:
\begin{itemize}
\item expansion of the query with possible aliases for the given name (alias component)
\item retrieval of documents containing mentions of the entity (information retrieval component
and entity linking component)
\item retrieval of sentences with mentions of the entity and possible slot fillers (candidate extraction component)
\item classification of the candidates (slot filler classification component)
\item postprocessing of the candidates (postprocessing component)
\end{itemize}
In the following section, our work on these different components is described in detail.

\section{Component description}
\subsection{Alias component}
\label{alias}
For query expansion, we used a pre-compiled list of possible aliases.
The aliases were obtained with JWPL~\cite{JWPL}, a Java-based Wikipedia interface.
For this study, we used a Wikipedia dump from July 2014 and extracted all redirect information.
In order to avoid noisy aliases, we implemented some basic cleaning steps like
minimum length of aliases or no aliases with another named entity type as the given entity.
For organizations, we also added various company-specific suffixes, such as ``Corp'', ``Co'', ``Inc''.
For persons, we included nicknames taken from the web\footnote{http://usefulenglishru/vocabulary/mensnames, 
http://usefulenglishru/vocabulary/womensnames} into the query expansion process.

Like last year, we used only the one alias with the lowest Levenshtein distance \cite{Levenshtein}
to the name given in the query
for the information retrieval component (IR alias). With this, we can cover spelling variations 
but reduce the number of falsly retrieved documents.
For the candidate extraction module, however, we used the whole list of aliases
to find as many occurrences of the entity as possible.

\subsection{Information retrieval component}
\label{IR}
For document retrieval, we used the open source system Terrier~\cite{Terrier}. 
We indexed the evaluation documents for the cold start task after some basic cleaning steps.
To retrieve documents relevant to a given person or organization, the following queries were used:
\begin{itemize}
 \item AND combination of the elements of the given name
 \item AND combination of the elements of the IR alias (see Section~\ref{alias})
 \item OR combination of the elements of the given name
\end{itemize}
We did not use phrase queries because we found that they did not
work well with spelling variations.
For geo-political entities, we only used the AND queries.
For each entity, we extracted up to 100 documents.

\subsection{Entity linking component}
\label{entityLinking}
The entity linking component was newly introduced into our 2015 system.
We did not use it for all runs since we wanted to investigate
its impact on the end-to-end performance.
The component used WAT \cite{WAT} to determine to which
Wikipedia entity the entity given by the query belongs to.
Then, for each document returned by the information retrieval document,
we checked whether the mention in the document refers to the
same Wikipedia entity as the query.
In case of a mismatch, the document was ignored by the end-to-end system.

\subsection{Candidate extraction component}
\label{candidateExtraction}
To find sentences with the entity in the retrieved documents, we applied fuzzy string matching 
(based on Levenshtein distance) and automatic coreference resolution. For coreference resolution,
we used Stanford CoreNLP \cite{coreNLP}. More details and analysis on this 
topic are presented in Section \ref{corefSec}.

After extracting sentences with mentions of the given entity, 
the system looked for possible fillers for the slot from the query. 
Similar to last year, we applied named entity recognition (with CoreNLP)
and a manual mapping from slots to possible named entity types of their fillers.
For string slots like per:title or per:charges, we assembled lists of possible filler
values based on Freebase \cite{Freebase}. In difference to last year, we 
used larger lists and also performed manual cleaning steps to improve their precision.

Furthermore, we immediately filtered impossible filler candidates like
floating point answers for number of employees of a company
or age of a person.

In difference to last year, our candidate extraction module has a recall
of 55\% to 62\% on the 2013 and 2014 evaluation data. Hence, its performance
has been doubled by keeping the number of false positive extractions almost constant.

\paragraph{Genre-specific document processing.}
The TAC 2015 evaluation corpus consists of news and discussion forum documents.
Those genres have different characteristics. Thus, it is reasonable to
process them in different ways. In our system, we applied special 
steps to discussion forum documents, such as ignoring 
text inside \texttt{<quote>} tags, normalizing casing of strings 
(e.g. mapping ``sErVice'' to ``service''), and using 
another flag for the sentence splitting component of Stanford CoreNLP.

\subsection{Slot filler classification component}
\label{classificationComponent}
In this evaluation, we used a variety of classifiers to decide
whether an extracted filler candidate is a valid filler for the given slot.
In particular, we used the distant supervised patterns by \cite{roth2013}, 
and trained support vector machines (SVMs)
with the same features as in our last year's system \cite{cis2014} as well as
two neural networks: a convolutional neural network and a recurrent neural network.
The classification component applied all these models to
score the context of a given entity - filler candidate pair. Their scores
were then combined by linear interpolation. The interpolation weights were
tuned based on previous TAC evaluation data.

\paragraph{Training data creation.}
For the SVM and the neural networks, we created a larger set of 
training examples compared to last year. We used distant supervision
with Freebase relation instances \cite{Freebase} and the following corpora:
\begin{itemize}
 \item TAC source corpus (LDC2013E45)
 \item NYT corpus (LDC2008T19)
 \item subset of ClueWeb\footnote{\url{http://lemurproject.org/clueweb12}}
 \item Wikipedia
 \item Freebase description fields
\end{itemize}
Negative examples were created in the same way as last year (by 
extracting sentences with entity
pairs with the correct named entity tags for the given slot that do not hold the 
negative relation according to Freebase). However, we also cleaned them with
trigger words and patterns: If a trigger/pattern of the negative relation appears in the
sentence, we would not include it into the set of negative examples.

\begin{figure}
\centering
 \includegraphics[width=.4\textwidth]{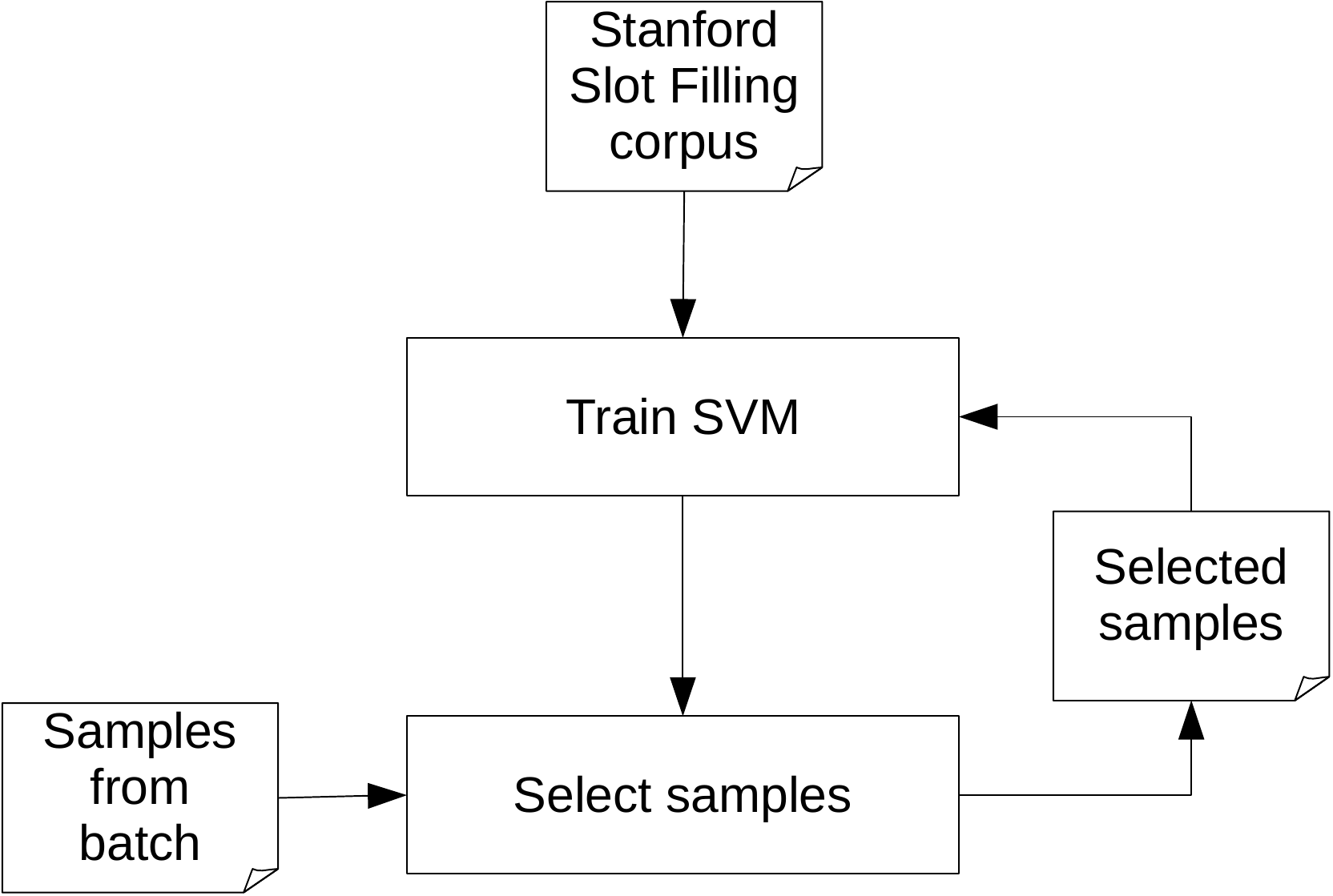}
 \caption{Training data selection process}
 \label{dataSelection}
\end{figure}

\paragraph{Training data selection.}
With this data creation process, we extracted a huge amount of data with noisy labels. 
To reduce the number of wrong labels, we performed an 
automatic training data selection process.
First, we divided the extracted training samples into $k$ batches.
Then, we trained one SVM per slot on the annotated slot filling data which
was released by Stanford last year \cite{active}.
Thus, the classifiers have been trained on data with presumably correct
labels and should, therefore, be able to help in the process
of selecting additional data.
For each batch of training samples, we used the classifiers to
predict labels for the samples and selected those samples for which the
distant supervised label corresponded to the predicted label and the
classifier had a high confidence.
Those samples were, then, added to the training data of the SVMs and
the SVMs were re-trained to predict the labels for the next batch.
This process is depicted in Figure \ref{dataSelection}.
The obtained training data set was then used to train the different classifiers.

\paragraph{Classification models.}
Based on the resulting data, one SVM, CNN and RNN was trained per slot.
The neural networks are described in detail in Section~\ref{neuralNets}.
To reduce redundant training, we trained only one classifier
for a slot and its inverse. 
We also merged the ``city'', ``country'' and ``stateorprovince'' slots 
to one ``location'' slot since we expect their fillers to appear in the same contexts.

Although we used many different sources to create training data, there were still slots for which not enough training data
could be extracted. Hence, no classifiers could be trained for them. Those slots were: per:charges, 
per:other\_family, per:religion, org:date\_dissolved, org:number\_of\_employees\_members,
org:political\_religious\_affiliation, org:shareholders. For them, we only used pattern matching in
the classification module.

\paragraph{Development data.}
In order to optimize the parameters of the models on data which is as clean as possible,
we automatically extracted sentences which correspond to the manually labeled
system outputs from the previous slot filling evaluations. Due to differences in the offset
calculation of some systems, not all available data could be used but the resulting
development data set still has a reasonable number of examples with presumably clean labels.
For more details on the development data and a script to reproduce the data, see \cite{adelNaacl2016}.

\subsection{Postprocessing component}
Finally, the classification component results were postprocessed. 
This included the following steps.
\paragraph{Output thresholds.}
Filler candidates with a classification score below a certain threshold were
discarded. The thresholds are slot-specific and have been tuned automatically
on previous evaluation data. 
This had outperformed 
slot independent thresholds in our last year's system.
For hop1 of one-hop queries, we 
increased the thresholds by 0.1 in order to reduce the number of false positive answers.
\paragraph{Location disambiguation.}
As mentioned in Section \ref{classificationComponent}, we did not distinguish between cities,
states or provinces, and countries in the classification component.
Before outputting the results, however, the extracted locations needed to be
disambiguated. The system decided based on city-, state- and country 
lists\footnote{\url{http://www.listofcountriesoftheworld.com},
 \url{http://en.wikipedia.org/wiki/List_of_U.S._state_abbreviations}, Freebase}
whether the location was a city, a state or province or a country.
\paragraph{Location inference.}
Based on city-to-state, city-to-country and state-to-country
mappings extracted from Freebase, we performed location inference for the case
that our system found a city or state while the given slot was a state or country.
\paragraph{Date normalization.}
For date slots, the extracted fillers were normalized to the output format (YYYY-MM-DD).
\paragraph{Filler candidate ranking.}
The extracted filler candidates were ranked according to their classification score.
For single-valued slots, only the top filler candidate was output. For list-valued
slots, the top $N$ filler candidates were output. ($N$ is
 slot-dependent and has been tuned on previous evaluation data in order to increase
 the precision of the system.)

\section{Coreference resolution for slot filling}
\label{corefSec}
 The importance of coreference resolution for slot filling has
 been shown before \cite{analysis2012,analysisRecall}. 
 Prior to the 2015 evaluations, we have investigated
 several aspects of coreference resolution in detail.
 We found several common errors of automatic coreference resolution
 that affect the end-to-end performance of the slot filling system.
 These errors include wrongly linked pronoun chains (pronouns linked to
 the wrong entity), unlinked pronoun chains
 (chains consisting of only pronouns) and no recognition of nominal
 anaphora (e.g., phrases like ``the 30-year-old'' are usually not recognized
 as being coreferent to an entity).
 For the last class of errors, we have developed a heuristic to deal with them:
 If the entity from the query occurs in sentence $t$ and sentence $t+1$ starts
 with a phrase like ``the XX-year-old'', ``the XX-based company'', ``the XX-born''
 and this phrase is not followed by another entity,
 there is a high chance that this phrase is coreferent to the entity.
 
 In order to reduce the runtime of our slot filling system, we pre-processed the
 TAC source corpus and 2015 evaluation corpus with coreference information.
 We have not processed all documents of the source corpus yet but so far 
 we have extracted about 36M coreference chains with a total number of 126M mentions.
 This resource will be publicly available to the 
 community.\footnote{We will provide it upon request by email.}

 In contrast to last year, we did not only use coreference information for 
 the entities from the queries but also for the fillers if the filler type
 was a person. Especially due to the newly introduced inverse slots, this turned out
 to improve the recall of the system considerably (e.g. consider the slot org:students and 
 the sentence ``He went to University of Munich.'')
 
 In end-to-end experiments, we have found that 
 the slot filling system is able to extract up to 12\% more true positive slot 
 fillers if it uses coreference resolution.
 While it also finds more false positive slot filler candidates
 in the candidate extraction step, almost all of these are ruled out
 by classification. Hence, coreference resolution turned out to be a 
 very important component in our slot filling system. In contrast to
 last year, we did not submit a run dedicated to coreference resolution
 this time. However, we ran our system 
 without the coreference resolution component after the official evaluation 
 to analyze its effect
 on the 2015 evaluation data. We report the results in Section \ref{analysis}.

\section{Neural networks for slot filling}
\label{neuralNets}
This section describes the neural networks which we trained
to extend our candidate classification component.
All of them used word embeddings to represent the words
in the input sentence. The embeddings have been trained with 
word2vec \cite{word2vec} on English Wikipedia.

\subsection{Convolutional neural networks}
Convolutional neural networks (CNNs) have been applied successfully to natural language processing
\cite{cw,kalchbrenner} in general and relation classification \cite{zeng,dosSantos} in particular.
We propose to also integrate them into an end-to-end slot filling system.
In contrast to prior work, we trained them on noisy distant supervised training data.
Our results show that they were still able to learn meaningful sentence representations.

\begin{figure}
\centering
\includegraphics[width=0.4\textwidth]{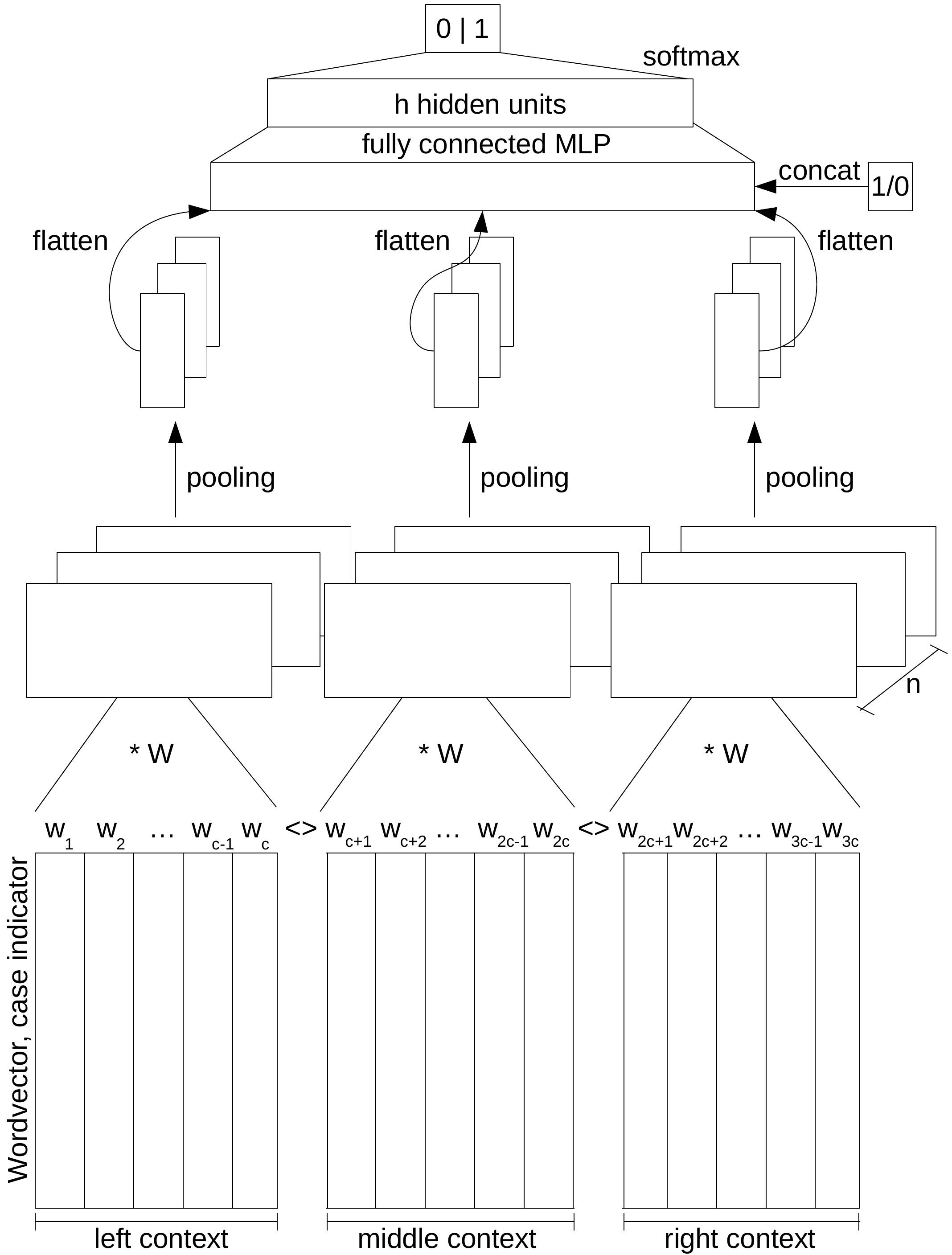}
\caption{Convolutional neural network for slot filling}
\label{cnnFig}
\end{figure}

CNNs are promising models for slot filler candidate classification out of two reasons:
(i) they create sentence representations and extract n-gram based features independent
of the position in the sentence, (ii) they use word embeddings as input and, thus, 
are able to recognize similar words or phrases (which are expected to have similar vectors).

For classification, we split the input sentence into three parts:
(1) the context left of entity and filler candidate, (2) the context
between entity and filler candidate, (3) the context right of
entity and filler candidate. Convolution and max pooling were applied to each of these
three parts individually. The weights for convolution, however, were shared to be able
to recognize relevant n-grams independent of their position in the input sentence.
Afterwards, the results were concatenated to one large vector. This vector
was extended with a flag indicating whether the entity or the filler candidate appeared 
first in the sentence. Then, it was passed to a
multi-layer perceptron consisting of a hidden layer and a softmax layer for classification.
The output of the network was binary: 1 if the context represented the given slot and
0 if it did not.

Figure \ref{cnnFig} depicts the structure of the CNN.

\subsection{Recurrent neural networks}
\label{rnnSec}
\begin{figure}
\centering
 \includegraphics[width=.45\textwidth]{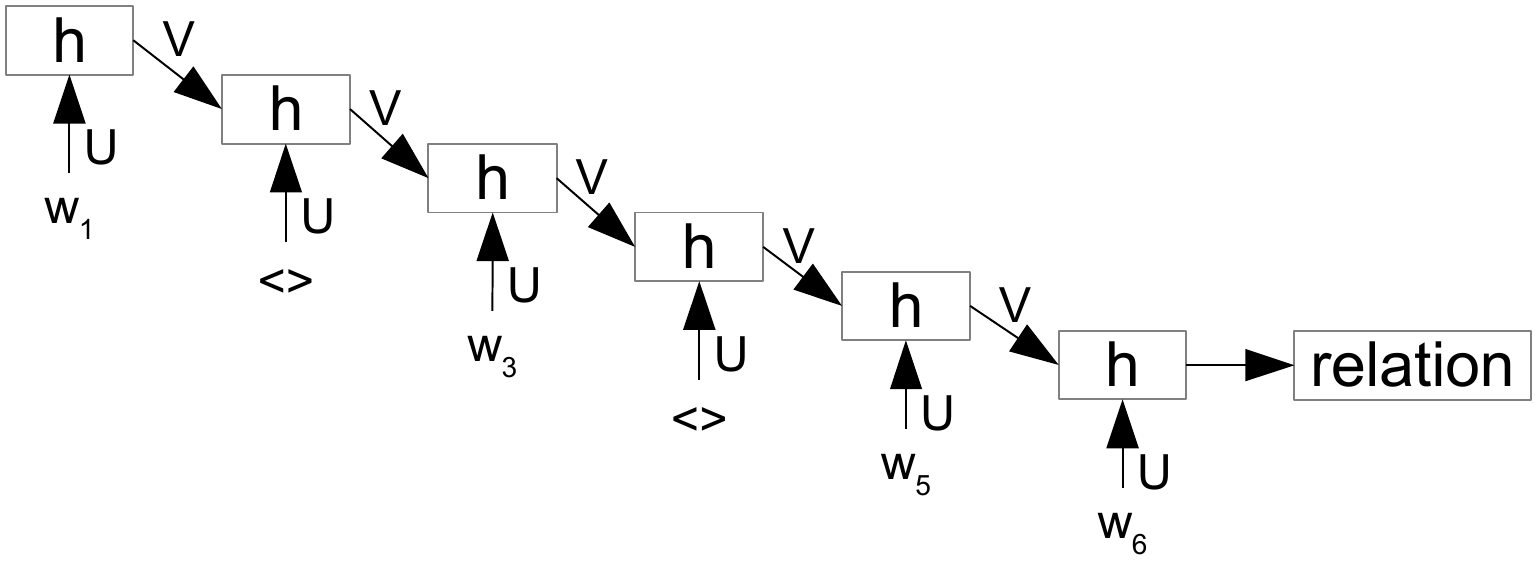}
 \caption{Uni-directional RNN for slot filling}
\label{rnnFig1}
\end{figure}

\begin{figure}
\centering
 \includegraphics[width=.5\textwidth]{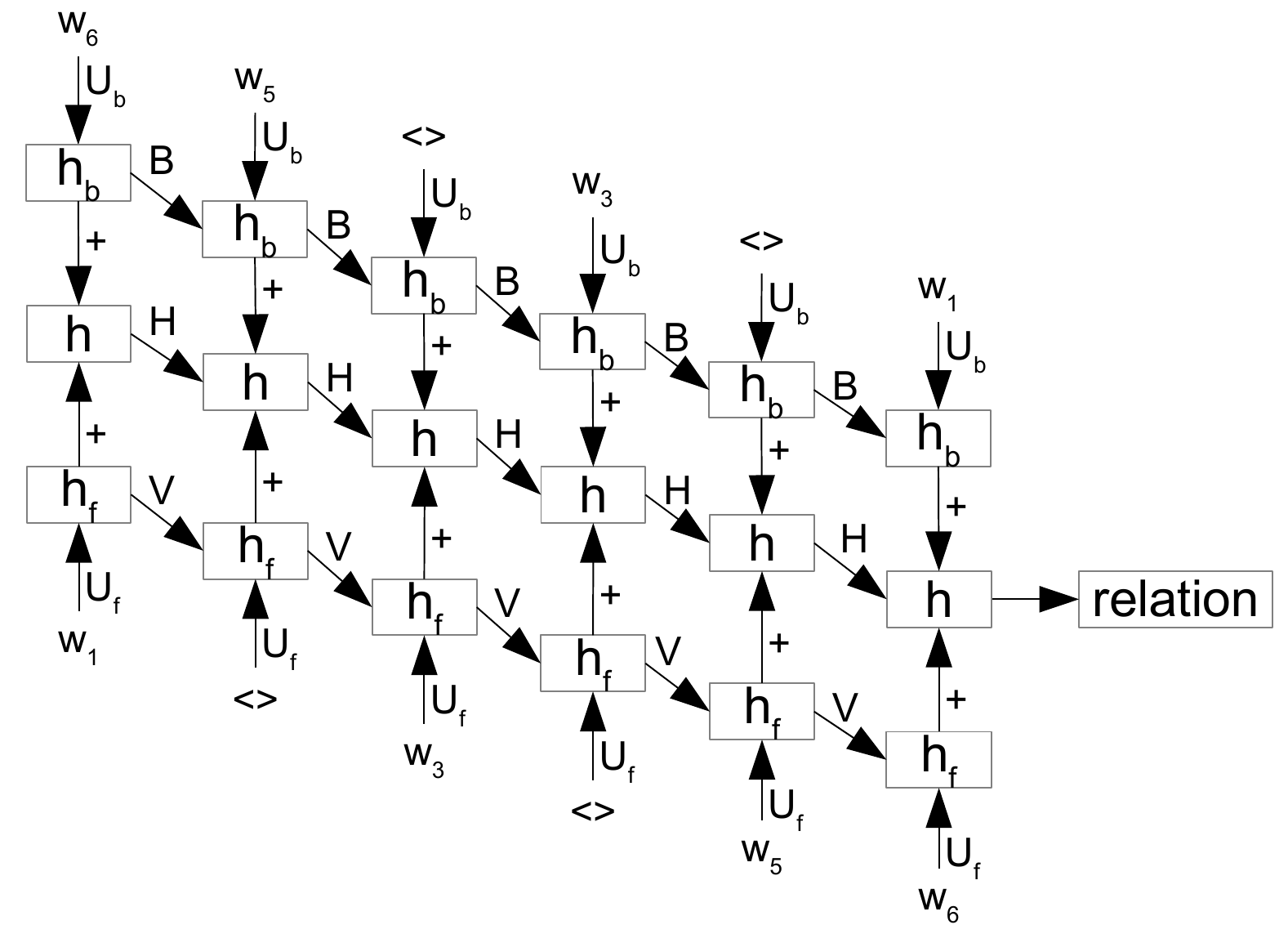}
 \caption{Bi-directional RNN for slot filling}
\label{rnnFig2}
\end{figure}
Recurrent neural networks (RNNs) have been applied successfully to
language modeling \cite{mikolovRNN}. \newcite{mvrnn} used recursive neural
networks based on dependency parse trees for relation classification.
In this work, we have integrated RNNs into our slot filling system.
In particular, we trained three different types of RNNs:
(1) a traditional forward RNN, (2) a bi-directional RNN, (3) a bi-directional
RNN trained in a multi-task fashion.
All RNNs first processed the whole sentence word by word and 
performed a relation classification step with a softmax layer afterwards.
The forward RNN processed the sentence only once, accumulated the
features of the input (represented by word embeddings) in its hidden layer
and predicted whether the input sequence was valid for the given slot.
The bi-directional RNN processed the sentence twice: from word 1 to word
n as well as from word n to word 1. It calculuated hidden layers for
both directions and accumulated them by summing their
values for the final prediction. The multi-task RNN predicted
the type of the next word (first relation argument, second relation argument or
other) at each time step and used this predicted type as an additional input
for the next word. 
In the slot filling system, we evaluated all three RNNs and took the 
decision of the most confident RNN as the final score.

Figures \ref{rnnFig1} and \ref{rnnFig2} depict the structures of the RNNs.

\section{Slot filling evaluation}
\subsection{Submitted runs}
\begin{table}
\centering
\begin{tabular}{rl|r|r|r}
& & P & R & F1\\
\hline
hop 0 & run 1 & \textbf{57.60} & 12.85 & 21.02\\
hop 0 & run 2 & 31.67 & 23.97 & 27.29\\
hop 0 & run 3 & 29.87 & \textbf{26.50} & \textbf{28.08}\\
hop 0 & run 4 & 31.71 & 24.13 & 27.41\\
hop 0 & run 5 & 19.11 & 22.32 & 20.59\\
\hline
hop 1 & run 1 & \textbf{15.89} & 1.89 & 3.38\\
hop 1 & run 2 & 10.46 & 6.33 & 7.89\\
hop 1 & run 3 & 14.13 & 5.89 & 8.31\\
hop 1 & run 4 & 11.82 & \textbf{7.00} & \textbf{8.79}\\
hop 1 & run 5 & 5.08 & 4.11 & 4.54\\
\hline
all & run 1 & \textbf{46.15} & 8.30 & 14.07\\
all & run 2 & 23.99 & 16.65 & 19.66\\
all & run 3 & 25.93 & \textbf{17.94} & \textbf{21.21}\\
all & run 4 & 24.63 & 17.02 & 20.13\\
all & run 5 & 14.48 & 14.76 & 14.62
\end{tabular}
\caption{End-to-end results, CSLDC max micro}
\label{results-ldcMax}
\end{table}

%%%% results from February 2016
% hop 0 & run 1 & \textbf{56.99} & 12.85 & 20.98\\
% hop 0 & run 2 & 31.83 & 23.97 & 27.35\\
% hop 0 & run 3 & 29.89 & \textbf{26.50} & \textbf{28.09}\\
% hop 0 & run 4 & 32.32 & 24.21 & 27.68\\
% hop 0 & run 5 & 19.19 & 22.32 & 20.63\\
% \hline
% hop 1 & run 1 & \textbf{20.24} & 2.48 & 4.42\\
% hop 1 & run 2 & 12.62 & 7.87 & 9.69\\
% hop 1 & run 3 & 14.41 & 7.43 & 9.81\\
% hop 1 & run 4 & 13.99 & \textbf{8.75} & \textbf{10.76}\\
% hop 1 & run 5 & 6.63 & 5.46 & 5.99\\
% \hline
% all & run 1 & \textbf{48.65} & 9.21 & 15.49\\
% all & run 2 & 25.89 & 18.32 & 21.46\\
% all & run 3 & 26.18 & \textbf{19.81} & \textbf{22.55}\\
% all & run 4 & 26.61 & 18.78 & 22.02\\
% all & run 5 & 15.81 & 16.55 & 16.17
%%%%%

We submitted the following runs to the official evaluation 2015:

\textbf{Run 1: High precision.}
Similar to our second run, this run used patterns, SVMs and
CNNs for classifying the filler candidates.
However, it only reported answers with high confidences (it added
0.2 to the output thresholds).
It can, thus, be considered a high precision run.

\textbf{Run 2: Patterns + SVM + CNN.}
This run can be considered as our base run. All the other runs
added or omitted one feature compared to this run in order to directly assess its impact
on the end-to-end performance.
In this run, we used patterns, SVMs and CNNs in the classification module.

\textbf{Run 3: Patterns + SVM + CNN + RNN.}
This run added RNN models as described in Section \ref{rnnSec}
to the classification component.

\textbf{Run 4: Entity linking.}
In this run, we applied the same classification module as in run 2.
Additionally, we used entity linking and
only considered those documents for filler candidate extraction which
included mentions of the same entity as the entity from the query (see Section \ref{entityLinking}).

\textbf{Run 5: Patterns + SVM.}
In order to assess the effect of adding neural networks to the classification module,
we only used traditional classification methods in this run (patterns and SVMs).

\subsection{Results and analysis}
\label{analysis}
Table \ref{results-ldcMax} shows detailed results of our runs.
The performance trends of the different runs are similar across
both hops and their combination (``all''): Run 1 had the highest precision but lowest
recall, run 3 and 4 (with RNNs and entity linking, respectively) 
led to the best F1 score. 
Compared to other slot filling systems, 
run 3 achieved rank 3 (see Table \ref{comparison}).

\begin{table}
\centering
\begin{tabular}{r|l|r}
Rank & Team & F1\\
\hline
1 & Stanford & 31.06\\
2&  UGENT &  22.38\\
\emph{3}& \emph{CIS} &  \emph{21.21}\\
4& UMass & 17.20\\
5& UWashington & 16.44
\end{tabular}
\caption{End-to-end result (CSLDC max micro) compared to other slot filling teams}
\label{comparison}
\end{table}

In experiments on previous evaluation data (2013 and 2014, slot filling track),
entity linking led to recall losses due to wrong decisions of the entity linker.
We suspect that the superior performance in this evaluation could be explained
by a large amount of ambiguous entity names (larger than in previous evaluations).

The RNNs (run 3) added small but consistent improvements to the final performance.
%We have expected them to improve the final performance of the system even more.
%The reason could be that the RNN models have not been fully optimized 
%at the time of the official evaluation due to time reasons. 

It is important to note that the performance difference of run 5 (without neural networks)
to run 2 and run 3 (with neural networks) was quite large (about 6 F1 points). This
shows the impact of neural networks. They improved the relation classification
and, thus, the end-to-end performance a lot even though they had been trained 
on noisy (distant supervised) training data.

\paragraph{Impact of coreference.}
After the official submissions, we ran the base run of our system (run 2) 
again without coreference resolution in the candidate extraction step.
Table \ref{coref} shows the end-to-end results
when using coreference (``run 2'') and when omitting it (``- coref'').
The number of true positives was reduced considerably (from 361 to 321)
when the system did not use coreference information.
The number of false positives was also lower, but the final results show
that the impact of the number of true positives was larger: The F1 scores
dropped by almost 5 points when omitting coreference resolution.

\begin{table}
\centering
\begin{tabular}{ll|r|r|r}
   & & P & R & F1\\
   \hline
hop 0 & run 2 & \textbf{31.67} & \textbf{23.97} & \textbf{27.29}\\
hop 0 & - coref & 19.33 & 22.40 & 20.75\\
\hline
hop 1 & run 2 & \textbf{10.46} & \textbf{6.33} & \textbf{7.89}\\
hop 1 & - coref & 5.32 & 4.11 & 4.64\\
\hline
all & run 2 & \textbf{23.99} & \textbf{16.65} & \textbf{19.66}\\
all & - coref & 14.83 & 14.81 & 14.82
   \end{tabular}
\caption{Impact of coreference resolution on end-to-end results, CSLDC max micro}
\label{coref}
\end{table}

\section{Conclusion}
This paper presented the CIS system for the TAC KBP Cold Start Slot Filling 
evaluation 2015. The system has been built upon our system from last year.
This paper showed the differences to our last year's system and paid
special attention to the classification and coreference module.
To improve the integration of coreference resolution, we have prepared a resource
and performed several analysis.
For the classification of slot filler candidates, we proposed to
use neural networks and showed that they improved end-to-end
performance by a large margin.
Our system achieved rank 3 of all slot filling systems in the 
official evaluations.

\section*{Acknowledgments}
Heike Adel is a recipient of the Google European Doctoral
Fellowship in Natural Language Processing and this
research is supported by this fellowship.
This work was also supported by DFG (grant
SCHU 2246/4-2).

We would like to thank Pankaj Gupta for 
his eager support with the RNN models.

\bibliographystyle{acl}
\bibliography{refs}

\begin{thebibliography}{}

\bibitem[\protect\citename{Adel and Sch\"{u}tze}2014]{cis2014}
Heike Adel and Hinrich Sch\"{u}tze.
\newblock 2014.
\newblock Tac kbp 2014 slot filling shared task: Baseline system for
  investigating coreference.
\newblock In {\em TAC}.

\bibitem[\protect\citename{Adel \bgroup et al.\egroup }2016]{adelNaacl2016}
Heike Adel, Benjamin Roth, and Hinrich Sch\"{u}tze.
\newblock 2016.
\newblock Comparing convolutional neural networks to traditional models for
  slot filling.
\newblock In {\em NAACL}.

\bibitem[\protect\citename{Angeli \bgroup et al.\egroup }2014]{active}
Gabor Angeli, Julie Tibshirani, Jean~Y. Wu, and Christopher~D. Manning.
\newblock 2014.
\newblock Combining distant and partial supervision for relation extraction.
\newblock In {\em EMNLP}.

\bibitem[\protect\citename{Bollacker \bgroup et al.\egroup }2008]{Freebase}
Kurt Bollacker, Colin Evans, Praveen Paritosh, Tim Sturge, and Jamie Taylor.
\newblock 2008.
\newblock Freebase: a collaboratively created graph database for structuring
  human knowledge.
\newblock In {\em SIGMOD international conference on Management of data}. ACM.

\bibitem[\protect\citename{Collobert \bgroup et al.\egroup }2011]{cw}
Ronan Collobert, Jason Weston, L{\'e}on Bottou, Michael Karlen, Koray
  Kavukcuoglu, and Pavel Kuksa.
\newblock 2011.
\newblock Natural language processing (almost) from scratch.
\newblock {\em JMLR}.

\bibitem[\protect\citename{Dos~Santos \bgroup et al.\egroup }2015]{dosSantos}
C{\'i}cero~Nogueira Dos~Santos, Bing Xiang, and Bowen Zhou.
\newblock 2015.
\newblock Classifying relations by ranking with convolutional neural networks.
\newblock In {\em ACL}.

\bibitem[\protect\citename{Ferschke \bgroup et al.\egroup }2011]{JWPL}
Oliver Ferschke, Torsten Zesch, and Iryna Gurevych.
\newblock 2011.
\newblock Wikipedia revision toolkit: Efficiently accessing wikipedia's edit
  history.
\newblock In {\em ACL-HLT System Demonstrations}. ACL.

\bibitem[\protect\citename{Kalchbrenner \bgroup et al.\egroup
  }2014]{kalchbrenner}
Nal Kalchbrenner, Edward Grefenstette, and Phil Blunsom.
\newblock 2014.
\newblock A convolutional neural network for modelling sentences.
\newblock In {\em ACL}.

\bibitem[\protect\citename{Levenshtein}1966]{Levenshtein}
Vladimir~I Levenshtein.
\newblock 1966.
\newblock Binary codes capable of correcting deletions, insertions and
  reversals.
\newblock In {\em Soviet physics doklady}.

\bibitem[\protect\citename{Manning \bgroup et al.\egroup }2014]{coreNLP}
Christopher~D. Manning, Mihai Surdeanu, John Bauer, Jenny Finkel, Steven~J.
  Bethard, and David McClosky.
\newblock 2014.
\newblock The {Stanford} {CoreNLP} natural language processing toolkit.
\newblock In {\em ACL System Demonstrations}.

\bibitem[\protect\citename{Mikolov \bgroup et al.\egroup }2011]{mikolovRNN}
Tomas Mikolov, Stefan Kombrink, Anoop Deoras, Lukar Burget, and Jan Cernocky.
\newblock 2011.
\newblock {RNNLM}-recurrent neural network language modeling toolkit.
\newblock In {\em ASRU Workshop}.

\bibitem[\protect\citename{Mikolov \bgroup et al.\egroup }2013]{word2vec}
Tomas Mikolov, Kai Chen, Greg Corrado, and Jeffrey Dean.
\newblock 2013.
\newblock Efficient estimation of word representations in vector space.
\newblock In {\em Workshop at ICLR}.

\bibitem[\protect\citename{Min and Grishman}2012]{analysis2012}
Bonan Min and Ralph Grishman.
\newblock 2012.
\newblock Challenges in the knowledge base population slot filling task.
\newblock In {\em LREC}.

\bibitem[\protect\citename{Ounis \bgroup et al.\egroup }2006]{Terrier}
Iadh Ounis, Gianni Amati, Vassilis Plachouras, Ben He, Craig Macdonald, and
  Christina Lioma.
\newblock 2006.
\newblock Terrier: A high performance and scalable information retrieval
  platform.
\newblock In {\em SIGIR Workshop on Open Source Information Retrieval (OSIR)}.
  ACM.

\bibitem[\protect\citename{Piccinno and Ferragina}2014]{WAT}
Francesco Piccinno and Paolo Ferragina.
\newblock 2014.
\newblock From tagme to wat: a new entity annotator.
\newblock In {\em First international workshop on Entity recognition \&
  disambiguation}.

\bibitem[\protect\citename{Pink \bgroup et al.\egroup }2014]{analysisRecall}
Glen Pink, Joel Nothman, and James~R Curran.
\newblock 2014.
\newblock Analysing recall loss in named entity slot filling.
\newblock In {\em EMNLP}.

\bibitem[\protect\citename{Roth \bgroup et al.\egroup }2013]{roth2013}
Benjamin Roth, Tassilo Barth, Michael Wiegand, Mittul Singh, and Dietrich
  Klakow.
\newblock 2013.
\newblock Effective slot filling based on shallow distant supervision methods.
\newblock In {\em TAC}.

\bibitem[\protect\citename{Socher \bgroup et al.\egroup }2012]{mvrnn}
Richard Socher, Brody Huval, Christopher~D Manning, and Andrew~Y Ng.
\newblock 2012.
\newblock Semantic compositionality through recursive matrix-vector spaces.
\newblock In {\em EMNLP-CoNLL}.

\bibitem[\protect\citename{Zeng \bgroup et al.\egroup }2014]{zeng}
Daojian Zeng, Kang Liu, Siwei Lai, Guangyou Zhou, and Jun Zhao.
\newblock 2014.
\newblock Relation classification via convolutional deep neural network.
\newblock In {\em COLING}.

\end{thebibliography}

% \begin{thebibliography}{}
% 
% \bibitem[\protect\citename{Aho and Ullman}1972]{Aho:72}
% Alfred~V. Aho and Jeffrey~D. Ullman.
% \newblock 1972.
% \newblock {\em The Theory of Parsing, Translation and Compiling}, volume~1.
% \newblock Prentice-{Hall}, Englewood Cliffs, NJ.
% 
% \bibitem[\protect\citename{{American Psychological Association}}1983]{APA:83}
% {American Psychological Association}.
% \newblock 1983.
% \newblock {\em Publications Manual}.
% \newblock American Psychological Association, Washington, DC.
% 
% \bibitem[\protect\citename{{Association for Computing Machinery}}1983]{ACM:83}
% {Association for Computing Machinery}.
% \newblock 1983.
% \newblock {\em Computing Reviews}, 24(11):503--512.
% 
% \bibitem[\protect\citename{Chandra \bgroup et al.\egroup }1981]{Chandra:81}
% Ashok~K. Chandra, Dexter~C. Kozen, and Larry~J. Stockmeyer.
% \newblock 1981.
% \newblock Alternation.
% \newblock {\em Journal of the Association for Computing Machinery},
%   28(1):114--133.
% 
% \bibitem[\protect\citename{Gusfield}1997]{Gusfield:97}
% Dan Gusfield.
% \newblock 1997.
% \newblock {\em Algorithms on Strings, Trees and Sequences}.
% \newblock Cambridge University Press, Cambridge, UK.
% 
% \end{thebibliography}

\end{document}